%% file: visionllama-main.tex
\documentclass[runningheads]{llncs}

\usepackage[final,year=2024,ID=8280]{eccv}

\usepackage{booktabs}
\usepackage{multicol,multirow}
\usepackage{makecell}
\usepackage{tabularx}
\usepackage{eccvabbrv}

\usepackage{graphicx}
\usepackage{booktabs}

\usepackage[accsupp]{axessibility}  %

\usepackage{bbding} 
\usepackage[weather]{ifsym}
\usepackage{fontawesome,eucal} 
\usepackage{colortbl}
\usepackage{pifont}

\usepackage{xr}
\makeatletter

\newcommand*{\addFileDependency}[1]{%
\typeout{(#1)}%
\@addtofilelist{#1}
\IfFileExists{#1}{}{\typeout{No file #1.}}
}\makeatother

\usepackage[pagebackref,breaklinks,colorlinks,citecolor=eccvblue]{hyperref}

\usepackage{colortbl}
\definecolor{mygray}{rgb}{0.89, 0.93, 0.85}
\definecolor{whitesmoke}{rgb}{0.96, 0.96, 0.96}
\definecolor{timberwolf}{rgb}{0.86, 0.84, 0.82}
\definecolor{baselinecolor}{gray}{.9}
\usepackage[capitalize]{cleveref}
\crefname{section}{Sec.}{Secs.}
\Crefname{section}{Section}{Sections}
\Crefname{table}{Table}{Tables}
\crefname{table}{Tab.}{Tabs.}

\usepackage{xspace}
\usepackage{array}
\newcolumntype{H}{>{\setbox0=\hbox\bgroup}c<{\egroup}@{}}
\newcommand{\ours}{VisionLLaMA\xspace}
\newcommand{\oursp}{Pyramid VisionLLaMA\xspace}

\newcommand{\ourdit}{DiT-LLaMA\xspace}
\newcommand{\oursit}{SiT-LLaMA\xspace}
\newlength\savewidth\newcommand\shline{\noalign{\global\savewidth\arrayrulewidth
  \global\arrayrulewidth 1pt}\hline\noalign{\global\arrayrulewidth\savewidth}}
  \newcolumntype{x}[1]{>{\centering\arraybackslash}p{#1pt}}

\def \altsmall   {Twins-SVT-S}
\def \altbase   {Twins-SVT-B}
\def \altlarge   {Twins-SVT-L}

\newcommand{\tabincell}[2]{\begin{tabular}{@{}#1@{}}#2\end{tabular}}
\newcommand{\baseline}[1]{\cellcolor{baselinecolor}{#1}}
\newcommand{\cmark}{\ding{51}\xspace}%
\newcommand{\xmarkg}{\textcolor{lightgray}{\ding{55}}\xspace}%

\def\T{{\!\top}}

\begin{document}

\title{VisionLLaMA: A Unified LLaMA Backbone for Vision Tasks
} 

\titlerunning{Vision LLaMA}

\author{
Xiangxiang Chu$^1$, ~~ Jianlin Su$^2$,  ~~ Bo Zhang$^1$,  ~~ Chunhua Shen$^3$
}

\authorrunning{Chu \it et al.}

\institute{Meituan Inc.%
\and 
Moonshot AI%
\and
Zhejiang University, China%
}


\maketitle

\begin{figure}
	\centering
    \includegraphics[width=.85\textwidth]{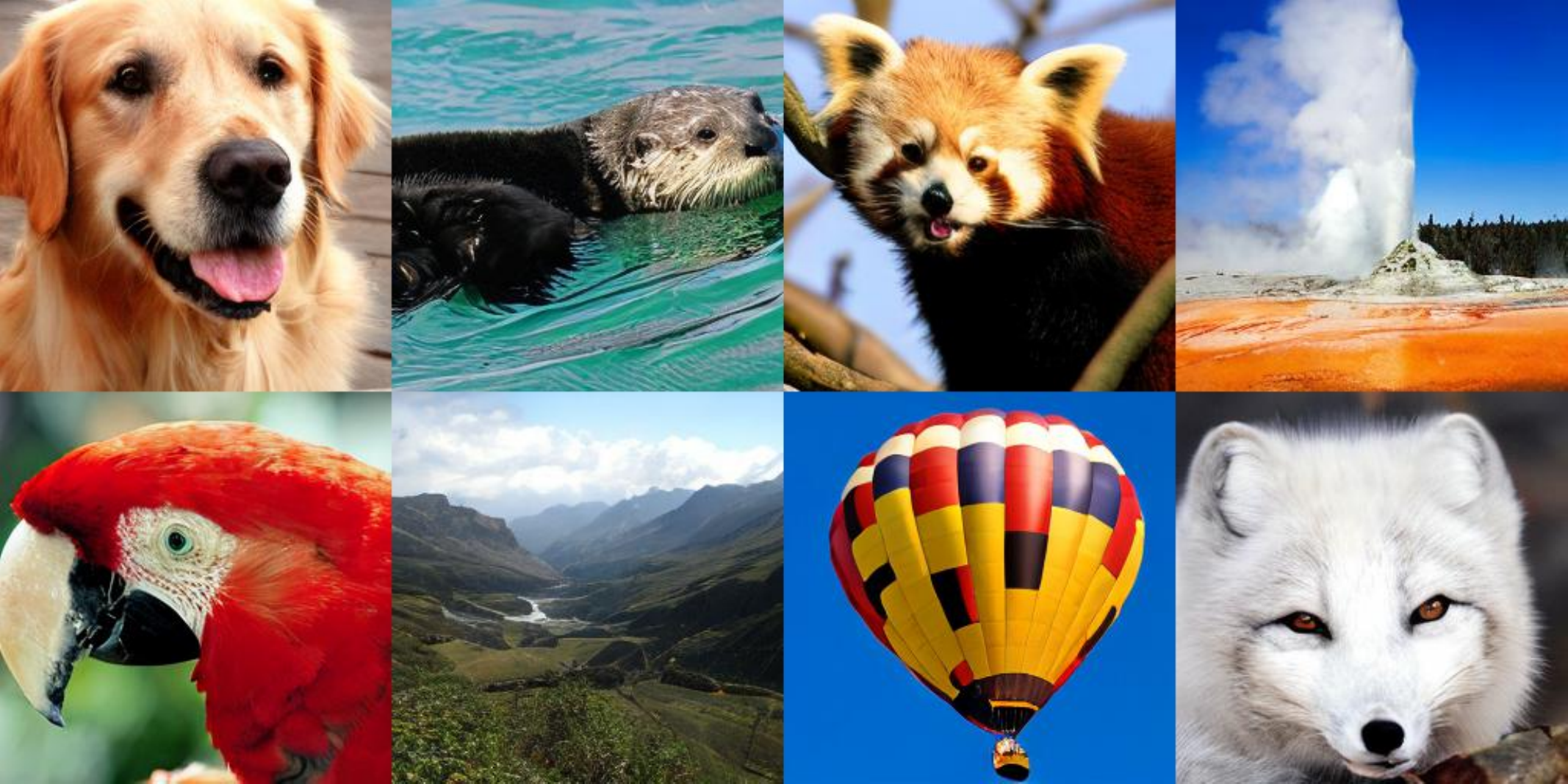}
    \caption{
    Generated images by \ourdit-XL of resolution (256, 256)  with CFG.}
	\label{fig:demo-img}
\end{figure}
\begin{abstract}
We all know that large language models are built on top of a transformer-based architecture to process textual inputs.  For example, the LLaMA family of models stands out among many open-source implementations. Can the same transformer be used to process 2D images? In this paper, we answer this question by unveiling a LLaMA-like vision transformer in plain and pyramid forms, %
termed 
{\bf \ours}, which is tailored for this purpose. 
\ours is a unified and generic modeling framework for solving most vision tasks.
We extensively evaluate its effectiveness using typical pre-training paradigms in a good portion of downstream tasks of image perception and especially image generation. In many cases, \ours 
has 
exhibited substantial gains over the previous state-of-the-art vision transformers. 
%
It is our
hope that researchers in computer vision can apply \ours
 presented here to solve various specific image generation and perception tasks.

Code is at: \url{https://github.com/Meituan-AutoML/VisionLLaMA}
\end{abstract}

\section{Introduction}\label{sec:intro}

Large language models have aroused great interest %
in the research 
community. One of the most influential and representative work is LLaMA \cite{touvron2023llama,touvron2023llama2}. Many recent works have converged to this architecture and solutions for various applications are built upon the open-sourced models. Besides, we have witnessed the blooming of multimodal models,
where many methods also heavily 
rely on LLaMA for text processing. Meanwhile, many endeavors~\cite{frantar2022gptq,xiao2023smoothquant,li2023norm} have been devoted to accelerating the inference speed and/or the memory cost of LLaMA. In a word, LLaMA is now the \textit{de facto}  architecture. 

Observing its success, a straightforward and interesting question is whether the LLaMA architecture can be another victory in the vision modality. If the answer is %
affirmative, 
then both vision and language models can use the same unified architecture and enjoy various deployment techniques designed for LLaMA on the fly. Unfortunately,  it is non-trivial to answer %
this question 
because there are some distinct differences between these two modalities. Firstly, it is common sense that text sequences are organized into one dimension, while vision requires two or more. Secondly, numerous vision tasks rely on pyramid backbones to perform better, while the LLaMA is a plain encoder. Thirdly, it is necessary to handle input images and videos with different resolutions. Our paper aims to resolve these difficulties and bridge the architectural gap between different modalities. 
Our main contributions are summarized as follows:
\begin{enumerate}
    \item We propose \ours, a vision transformer architecture similar to L\-L\-a\-M\-A to reduce the architectural differences between language and vision.
    \item 

We investigate means to adapt \ours to tackle common vision tasks, including image comprehension and creation (Figure~\ref{fig:demo-img}). We examine two well-known vision architecture schemes (plain and pyramid) and assess their performance under supervised and self-supervised learning scenarios. Additionally, we introduce AS2DRoPE (\ie auto-scaled 2D RoPE), which expands rotated positional encoding from 1D to 2D and utilizes interpolation scaling to accommodate arbitrary resolutions.

    \item Without bells and whistles, \ours significantly outperforms the wi\-de\-sp\-r\-ead and carefully fine-tuned vision transformer by clear margins across many representative tasks such as image generation, classification, semantic segmentation, and object detection. Extensive experiments indicate that \ours demonstrates faster convergence speed and better performance than existing vision transformers. 
    
\end{enumerate}

\section{Related Work}
\label{sec:related}

\par \noindent \textbf{Vision Transformer.} ViT~\cite{dosovitskiy2020image} successfully applied Transformer~\cite{vaswani2017attention} from natural language processing to the vision world and many more efficient and powerful follow-up works are induced, like  DeiT~\cite{touvron2022deit}, Swin~\cite{liu2021swin}, PVT~\cite{wang2021pyramid}, and Twins~\cite{chu2021Twins}. The pre-training paradigm has been shifted from supervised learning on large-scale categorically labeled datasets like ImageNet~\cite{deng2009imagenet} to unsupervised learning~\cite{he2022masked}.  DiT~\cite{peebles2023scalable} adopts a transformer that operates on latent patches for diffusion models~\cite{sohl2015deep,ho2020denoising}, outperforming the commonly used U-Net backbone~\cite{ronneberger2015u}.

\par \noindent \textbf{Large Language/Multi-modal  Models} Proprietary models like GPT4~\cite{gpt4} have been taking the lead in the LLM competition, though their technical details are hidden from the public. In contrast, the community has blossomed to release a myriad of open-source counterparts. For instance, BLOOM~\cite{scao2022bloom} and LLaMA~\cite{touvron2023llama} catch up with the performance of the closed model GPT-3~\cite{brown2020language}. Later in copious detail, LLaMA-2~\cite{touvron2023llama2} describes a pack of architectural tweakings including pre-normalization called RMSNorm~\cite{zhang2019root}, the activation function SwiGLU~\cite{shazeer2020glu}, rotary positional embeddings RoPE~\cite{su2023roformer}, as well as a dedicated training pipeline, which comprises self-supervised pre-training and supervised fine-tuning enhanced by Reinforcement Learning with Human Feedback (RLHF). Many vision language models \cite{liu2023llava,liu2023improved, zhu2024minigpt,lai2023lisa, wei2023lenna} are built on LLaMA and show impressive results on the visual dialog, reasoning, perception, and so on. The LLaMA architecture has also been applied in resource-limited multimodal scenarios such as mobile phones \cite{chu2023mobilevlm,chu2024mobilevlm} recently and shows potential applications.

 \par \noindent \textbf{Positional Encoding for Transformers.} Transformer~\cite{vaswani2017attention} originally comes with 2D absolute position embeddings in sinusoidal forms. In contrast, the relative ones as in ~\cite{shaw2018self} pay attention to the relations of input tokens and can handle variable lengths of sequences. Rotary positional embeddings~\cite{su2023roformer} are introduced to encode both absolute and relative positional information, which is proven to be effective in large language models~\cite{touvron2023llama}. Conditional positional embeddings ~\cite{chu2023conditional} are proposed to add positional information for vision transformers according to the input image, with the benefit of boosted performance and generalizability to arbitrary input resolutions. As for LLMs, the models are usually pre-trained with a given fixed context length \cite{touvron2023llama, touvron2023llama2,yang2023baichuan} and then fine-tuned to a larger context length to support long context inference. \cite{chen2023extending} extends the context length of LLaMA by simple positional interpolations. Base frequency adjustment of RoPE is also studied by \cite{xiong2023effective} to enable long-context continued training. NTK-Aware scaled RoPE allows LLaMA to have an extended context size without fine-tuning and minimal perplexity degradation \cite{roziere2023code}.

\section{Method}
\label{sec:method}

\subsection{Plain Transformer}

\begin{figure}[t!]
	\centering
	\includegraphics[width=0.85\columnwidth]{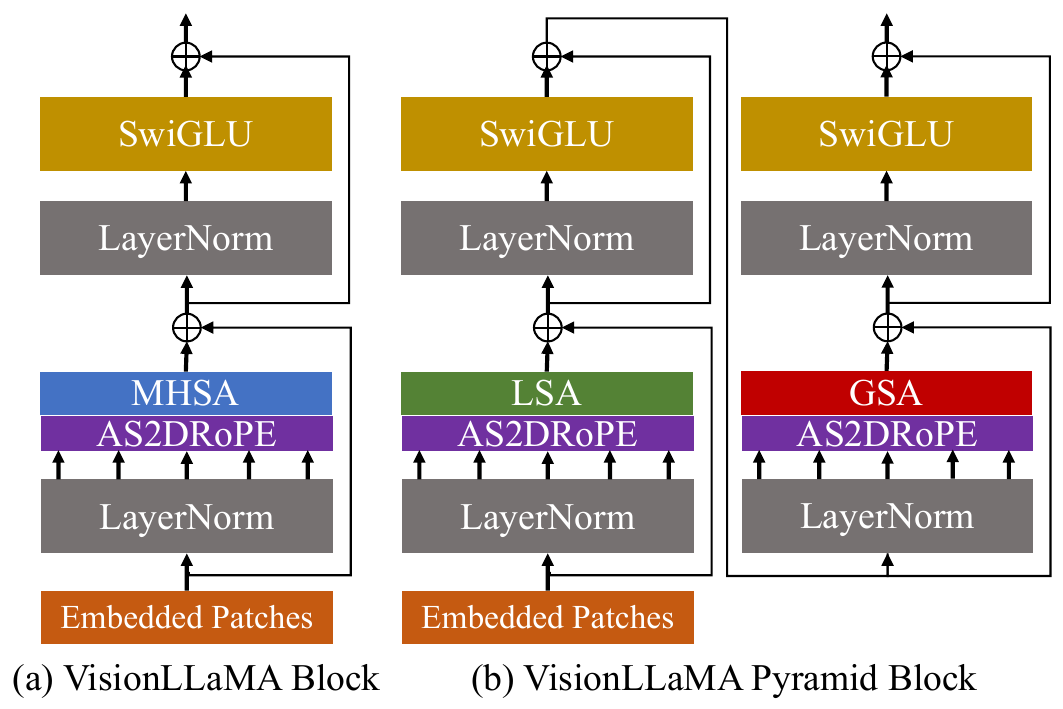}
	\caption{\ours block (a) in plain Transformer and (b) in pyramid Transformer.}
	\label{fig:transformer-block}
\end{figure}

Our plain \ours follows the pipeline of ViT \cite{dosovitskiy2020image} and we retain the architecture design of LLaMA as closely as possible. For an image of $H\times W$, it's firstly transformed and flattened into $N=\frac{H \times W}{P^2}$ non-overlapped patches $X\in {\cal R}^{N \times C}$. Then a class token is prepended at the beginning of the sequence and the whole sequence is processed by $L$ \ours blocks. Unlike \cite{dosovitskiy2020image}, we do not add positional encodings to the input sequence since our basic block readily contains positional encoding. Specifically, the basic block differs from the standard ViT block by two components: self-attention with positional encoding (RoPE) \cite{su2023roformer} and SwiGLU activation \cite{shazeer2020glu}. We still utilize LayerNorm \cite{ba2016layer} instead of RMSNorm \cite{zhang2019root} since we find the former behave better through the classification experiment (see Table~\ref{tab: ablation_norm}). The basic block is illustrated in Figure~\ref{fig:transformer-block} (a). It should be noted that directly applying 1D RoPE in vision tasks cannot well generalize to other resolutions, which is different from the training resolution. Therefore, we extend it to the 2D form. It can be formally written as,
\begin{equation}
	\begin{split}
	&{{\bf{z}}^{l}_{ij}} = \text{MHSA}\left(\text{AS2DRoPE} \left({\text{LayerNorm}\left( {{{\bf{z}}^{l - 1}_{ij}}} \right)} \right)\right) + {\bf{z}}^{l - 1}_{ij}, \\
&{{\bf{z}}^l_{ij}} = \text{SwiGLU}\left( {\text{LayerNorm}\left( {{{\bf{z}}^{l}}_{ij}} \right)} \right) + {{\bf{z}}^{l}_{ij}},\\
& i \in \{1, 2, ...., m\}, j \in \{1, 2, ...., n\}.
	\end{split}
 \label{eq:plain_vit}
\end{equation}
where ${z}^{l}_{ij}$ means the output of the $l$ block at position ($i,j$).

\subsection{Pyramid Transformer}
It is straightforward to apply \ours to window-based transformers that utilize additive relative position encoding, such as Swin~\cite{liu2021swin}. In this paper, we choose a stronger baseline Twins \cite{chu2021Twins} to explore how to build a powerful pyramid transformer under strictly controlled settings. The original architecture of Twins exploits a conditional position encoding and interleaved local-global information exchange in the form of local and global attention. These components can be found in various transformers, which means it is not difficult to apply \ours in other pyramid transformer variants by following our method. Note that our target is not to invent a novel pyramid vision transformer, but to show how we adapt the basic design of \ours based on the existing ones.  Therefore, we simply conform to the smallest modifications to the architecture and hyperparameters. Following the name convention of \cite{chu2021Twins}, the two consecutive blocks can be written as, 
\begin{equation}
	\begin{split}
	&{{\hat{\bf{z}}}^{l}_{ij}} = \text{LSA}\left(\text{AS2DRoPE} \left({\text{LayerNorm}\left( {{{\bf{z}}^{l - 1}_{ij}}} \right)} \right)\right) + {\bf{z}}^{l - 1}_{ij}, \\
&{{\bf{z}}^l_{ij}} = \text{SwiGLU}\left( {\text{LayerNorm}\left( {{{\hat{\bf{z}}}^{l}}_{ij}} \right)} \right) + {{\hat{\bf{z}}}^{l}_{ij}},\\
&{{\hat{\bf{z}}}^{l+1}} = \text{GSA} \left(\text{AS2DRoPE} \left( {\text{LayerNorm}\left( {{{\bf{z}}^{l}}} \right)} \right)\right) + {\bf{z}}^{l}, \\
&{{\bf{z}}^{l+1}} = \text{SwiGLU}\left( {\text{LayerNorm}\left( {{{\hat{\bf{z}}}^{l+1}}} \right)} \right) + {{\hat{\bf{z}}}^{l+1}}, \\
& i \in \{1, 2, ...., m\}, j \in \{1, 2, ...., n\}. %
\label{eq:gvt}
	\end{split}
\end{equation}
 where LSA is the local self-attention operation within a group and GSA is the global sub-sampled attention by interacting with the representative keys from each sub-window $\hat{\bf{z}}_{ij} \in \mathcal{R}^{k_1\times k_2 \times C}$ and $m\times n$ is the sub-window shape. 

We remove the conditional position encoding in our pyramid \ours since AS2DRoPE already contains positional information. Besides, we also remove the class tokens and use GAP (global average pooling) before the classification head as \cite{chu2021Twins, chu2023conditional}. The basic block in this setting is illustrated in Figure~\ref{fig:transformer-block}(b).

\subsection{Training or 
Inference 
Beyond the Sequence Length}

\textbf{From 1D RoPE to 2D.} Handling different input resolutions is a common requirement in vision tasks. Convolutional neural networks use the sliding window mechanism to deal with the variable length. In contrast, most vision transformers apply local window operations or interpolations. For instance, DeiT \cite{touvron2022deit} adopts bicubic interpolations when trained on different resolutions. CPVT \cite{chu2023conditional} uses convolution-based position encoding. Here we evaluate the performance of 1D RoPE~\cite{su2023roformer}. Specifically, our pyramid \ours based on Twins-SVT-S with 1D RoPE achieves 81.5\% top-1 accuracy on an input of 224$\times$224. However, the performance severely degrades to zero when evaluated on 448$\times$448. Therefore, we extend the 1D RoPE to 2D. As for the multi-head self-attention, the 2D RoPE is shared across different heads. Specifically, given a token $x_{i,j} \in {\cal R}^{d}$, we obtain its position-encoded token $x^{\rm PE}_{i,j} = \textbf{R}_{i,j} x_{i,j}$, and the  diagonal matrix  $\textbf{R}_{i,j}  \in {\cal R}^{d\times d}$ can be written as, 
\[
\scriptsize
\resizebox{1\hsize}{!}{$
\begin{bmatrix}
    \cos(i\theta_0)       & -\sin{(i\theta_0)} & 0 &0& \dots & 0 &0 &0 \\
    \sin(i\theta_0)       & \cos{(i\theta_0)} &  0 &0& \dots &0&0&0\\
    0&0& \cos(j\theta_0)       & -\sin{(j\theta_0)} &\dots &0&0&0\\
    0&0&  \sin(j\theta_0)       & \cos{(j\theta_0)} &\dots &0&0&0\\
     \\
    0&0&0&\dots&\cos(i\theta_{d-4})       & -\sin{(i\theta_{d-4})} & 0 &0\\
    &&&&\sin(i\theta_{d-4})       & \cos{(i\theta_{d-4})} &  0 &0\\
    0&0&0&\dots&0& 0 &\cos(j\theta_{d-4})       & -\sin{(j\theta_{d-4}})  \\
    0&0&0&\dots&0&0&\cos(j\theta_{d-4})       &\cos{(j\theta_{d-4})}  \\
\end{bmatrix}
$
}
\]
where $\theta_m = 10000^{-m/d}$ and $m \in \{0, 4, 8, ..., d-4\} $. Note that $\textbf{R}$ is an orthogonal matrix. We make minor modifications to the frequency selection \cite{su2023roformer} and make two axes share the same frequency.  It is easy to verify that
\begin{equation}
    {\bf R}_{i_1,j_1}^{\T} 
    {\bf R}_{i_2,j_2} =  {\bf R}_{i_1-i_2,j_1-j_2}.
\end{equation}

\textbf{Positional interpolation helps 2D RoPE to better generalize.} Inspired by \cite{chen2023extending}, which uses interpolation to extend the context window of LLaMA, involving higher resolution is analogous to extending the 2D context window of \ours. Unlike the language task \cite{chen2023extending} with an enlarged fixed context length, vision tasks like object detection usually deal with different sampled resolutions at different iterations. We train our small model using an input resolution of 224$\times$224 and evaluate the performance on the larger resolutions without re-training, which guides us to apply good strategies of interpolation or extrapolation. Consequently, we apply \emph{auto-scaled interpolation} (so-called AS2DRoPE) based on an `anchor resolution'. Without loss of generality, we assume handling the square image of $H \times H$ and an anchor resolution  $B\times B$ during the training, we calculate  
\begin{equation}
    \mathbf{R^{\prime}}_{i,j} x_{i,j}=
    \mathbf{R}_{i \cdot B/H,j \cdot B/H},
\end{equation}
which can be efficiently implemented and does not introduce an extra cost.
Note if the training resolution is kept unchanged, AS2DRoPE degenerates as a 2D RoPE.

As for the GSA under the pyramid setting, we require special treatments since we need to add positional information to the summarized keys.  These sub-sampled keys are generated by abstraction on the feature maps. Without loss of generality, we use a convolution with a kernel size of $k\times k$ and stride of $k$. The coordinate of the generated key can be formulated as the average of the sampled features. We show a simple example in Figure~\ref{fig:gsa-encoding} (appendix.).

\section{Experiments}

We evaluate the effectiveness of \ours on image generation, classification, segmentation, and detection. Unless otherwise specified, all models are trained on 8 NVIDIA Tesla A100 GPUs.
\subsection{Image Generation}
\textbf{Image generation based on the DiT framework.} We apply \ours under the DiT framework \cite{peebles2023scalable}, which is a representative work of image generation using vision transformers and DDPM \cite{ho2020denoising}. Specifically, we replace the original vision transformer of DiT with \ours while keeping other components unchanged.  This controlled experiment manifests the generality of \ours on the image generation task. Moreover, we do not change the original hyper-parameters, although it may be sub-optimal to achieve the best performance. We also use the pre-trained VAE \cite{kingma2013auto} (the ft-EMA VAE model) from SD\cite{rombach2022high}, which has a down-sample factor of 8.  For classifier-free guidance, we use a coefficient of $1.5$.  The training resolution of the image is 256 $\times$ 256. As suggested by \cite{peebles2023scalable}, we choose the strongest adaLN-Zero version as our implementation.
We also use flash attention \cite{dao2023flashattention} and mixed precisions to speed up the training. Note that FID is known to be sensitive to small implementation details \cite{parmar2022aliased}. To make accurate calculations and fair comparisons, we use the TensorFlow tool from \cite{dhariwal2021diffusion} as \cite{peebles2023scalable}.

We choose 250 sample steps of DDPM as \cite{peebles2023scalable} and show the result in Table~\ref{tab:compare-with-DiTs}. As a common practice, FID is regarded as a primary metric. We also report other secondary metrics such as sFID \cite{nash2021generating}, Precision/Recall \cite{kynkaanniemi2019improved}, and Inception Score \cite{salimans2016improved}. Most experiments are controlled on 400k training steps. \ours significantly outperforms DiT across various model sizes. We also extend the training steps of XL models to 2352k steps to evaluate whether our models have the faster convergence advantage or still behave better under the setting of longer training epochs. \ourdit-XL/2 has 0.83 lower FID \cite{heusel2017gans} than DiT-XL/2, 
indicating that  \ours not only has better computing efficiency but higher performance than DiT. We show some generated samples %
in Figure~\ref{fig:demo-img} using our XL model.

\begin{table*}[ht]
\centering
\caption{Image generation comparisons using DiT \cite{peebles2023scalable}.}
\label{tab:compare-with-DiTs}
\setlength{\tabcolsep}{2pt}
\scriptsize 
\begin{tabular}{*{1}{l}*{2}{c}*{10}{c}}
\toprule
Model & CFG & Flops & Params & Steps & lr & FID%
$\downarrow$& sFID$\downarrow$ &Precision$\uparrow$ &Recall$\uparrow$ &IS$\uparrow$\\
& & (G) & (M) & (K) & & & & \\
\midrule
DiT-B/4 & N & 5.56 & 130 & 400 & 0.0001& 68.38& 12.66&36.07 &54.71&20.27\\
\rowcolor{yellow!50} \ourdit-B/4 &N & 5.56 & 130 & 400& 0.0001&63.17&12.63&38.27&56.75&22.47\\
DiT-B/4&Y&5.56&130&400&0.0001&45.38&9.97&46.89&53.66&34.27\\
\rowcolor{yellow!50} \ourdit-B/4 &Y & 5.56 & 130 & 400& 0.0001&39.51&9.82&50.46&54.75&40.17\\
\midrule
DiT-L/4 & N &19.70 & 458 & 400& 0.0001& 44.37 &8.97 &48.16&61.53&32.25\\
\rowcolor{yellow!50} \ourdit-L/4 & N &19.70 & 458 & 400& 0.0001& 40.32&9.04&49.87&61.61&36.56 \\
DiT-L/4 & Y &19.70 & 458 & 400& 0.0001& 22.51&7.08&62.67&55.27&66.58\\
\rowcolor{yellow!50} \ourdit-L/4 & Y &19.70 & 458 & 400& 0.0001&18.64&7.01&65.40&54.35&78.52\\
\midrule
DiT-XL/4 &N & 29.05&675& 400& 0.0001 & 43.01 &-&-&-& -\\
\rowcolor{yellow!50} \ourdit-XL/4 & N &29.05&675& 400& 0.0001 & 35.99& 8.48 & 52.31 & 61.65&41.18 \\
DiT-XL/4 &Y & 29.05&675& 400& 0.0001 & 22.52 &7.09&62.68&55.27&66.58 \\
\rowcolor{yellow!50} \ourdit-XL/4 &Y &29.05&675& 400& 0.0001&18.69&7.02&65.67&55.57&78.32\\
\midrule
DiT-XL/2 & N &118.64& 675 &2352& 0.0001 &10.67 & -&-&-&-\\
\rowcolor{yellow!50} \ourdit-XL/2 & N &118.64& 675 &2352& 0.0001 &9.84 & 6.47 & 67.45 &66.71&117.72\\
\rowcolor{yellow!50} \ourdit-XL/2 & Y &118.64& 675 &2352& 0.0001 & \textbf{2.42} & 4.51 & 83.03 &56.82&265.39\\

\bottomrule
\end{tabular}
\end{table*}

\subsection{Classification on ImageNet}

\subsubsection{Supervised Training}
\begin{table}
 \setlength\tabcolsep{1pt}
	\centering
	\scriptsize  	
    \caption{Comparisons on ImageNet-1K supervised classification. All the models are trained using the ImageNet-1K dataset. $\dagger$: retrained using the official code. 160I 800E+224I 20E means two-stage training, the model is firstly trained for 800 epochs using 160$\times$160, then trained for 20 epochs with higher image resolution 224$\times$224.}
   	\label{tab:classfication_sft_imagenet_1k}
	\begin{tabular}{rcHHcc}
		\toprule
		Model & Param & FLOPs & Throughput  & Setting& Top-1  \\
        & (M) & (G) & (Images$/$s) &  & (\%)  \\
		\midrule
		DeiT-Small\cite{pmlr-v139-touvron21a}  & 22 & 4.6 & &224I 300E& 79.9 \\
		CPVT-Small-GAP \cite{chu2023conditional} & 23 & 4.6& 817& 224I 300E & 81.5 \\
              DeiT3-Small \cite{touvron2022deit} &22&&&224I 800E&81.4	\\
\rowcolor{yellow!50}              \ours-S \cite{touvron2022deit} &22&&&224I 800E&81.6	\\

		Swin-T \cite{liu2021swin} & 29 & 4.5& 766 & 224I 300E &81.3 \\
		\altsmall \cite{chu2021Twins} & 24 &2.9& 1059 &224I 300E&81.7  \\
\rowcolor{yellow!50}            \oursp-S &24&&& 224I 300E & 81.6\\

		\midrule
		Swin-S~\cite{liu2021swin} & 50 & 8.7 & 444& 224I 300E &83.0\\
		\altbase \cite{chu2021Twins}  &56 &8.6& 469 & 224I 300E&83.2 \\
\rowcolor{yellow!50}            \oursp-B &56&&&224I 300E & 83.2\\
		\midrule
            DeiT3-Base \cite{touvron2022deit} &86 &&&192I 800E + 224I 20E&83.8\\
\rowcolor{yellow!50}            \ours-B &86&&&192I 800E + 224I 20E & 83.6\\
	    Swin-B \cite{liu2021swin} & 88 & 15.4& 275& 224I 300E &83.3 \\
		\altlarge \cite{chu2023conditional}   & 99& 15.1 & 288 &  224I 300E & 83.7 \\
\rowcolor{yellow!50}            \oursp-L& 99&&& 224I 300E&83.6\\
		\midrule
            DeiT3-Large$^\dagger$ &310&&&160I 800E+224I 20E&84.5\\
\rowcolor{yellow!50}            \ours-L  &310&&&160I 800E+224I 20E& \textbf{84.6}\\
		\bottomrule
	\end{tabular}
\end{table}

In this section, we focus on supervised training on the ImageNet-1K dataset~\cite{deng2009imagenet} to make fair comparisons. We exclude other datasets or distillation tricks. All the models are trained using the ImageNet-1K training set, and we report the accuracy of the validation set in  Table~\ref{tab:classfication_sft_imagenet_1k}.

\par \noindent\textbf{Plain Vision Transformer Comparison.} DeiT3~\cite{touvron2022deit} is the state-of-the-art plain vision transformer, which proposes special data augmentations and performs extensive hyperparameter search to boost the performance of DeiT \cite{pmlr-v139-touvron21a}. During the reproduction of DeiT3, we observe that it is sensitive to hyperparameters and prone to overfitting. Replacing the class token with  GAP (global average pooling)\cite{chu2023conditional} leads to a 0.7\% top-1 accuracy drop for the DeiT3-Large model after 800 epochs of training. Therefore, we use the class token instead of GAP in the plain transformer and report the result in Table~\ref{tab:classfication_sft_imagenet_1k}, where \ours achieves a top-1 accuracy comparable to DeiT3. The detailed hyperparameter is listed in the appendix. Note that the accuracy on a single resolution does not provide comprehensive comparisons, we also evaluate the performance across different image resolutions as \cite{chu2023conditional} and report the result in Table~\ref{fig: acc_vs_resolution}(supp.). As for DeiT3, we use the bicubic interpolation for the learnable positional encoding.  Although these two models have comparable performance at the resolution of 224$\times$224, the gap is enlarged when the resolution is increased, which means our method generalizes better across different resolutions, which is a vital function for many downstream tasks such as object detection.

\par \noindent \textbf{Pyramid Vision Transformer.}
We use the same architecture as Twins-SVT \cite{chu2021Twins} and the detailed configuration is listed in Table~\ref{tab:pyramid_arch} (supp.). We remove the conditional position encoding since \ours already contains one kind of rotary position encoding. Therefore, \ours is a convolution-free architecture. We do not tune the hyper-parameters and directly follow the setting provided in \cite{chu2021Twins}. Although it's suboptimal, it can still achieve competitive performance. As \cite{chu2021Twins, chu2023conditional}, we do not use the class token and apply GAP. The result is shown in Table \ref{tab:classfication_sft_imagenet_1k} and our method achieves comparable performance as Twins across various levels of models and outperforms Swin \cite{liu2021swin} consistently.  We further compare the pyramid transformers using popular downstream tasks, which are shown in the later sections.  

\subsubsection{Self-Supervised Training}
To make fair comparisons, we limit the training data to ImageNet-1K. We also exclude any component that utilizes  CLIP~\cite{radford2021learning}, DALLE~\cite{DALLE}, or distillation, which can be orthogonally combined to further boost the performance.  Our implementation is based on the MMPretrain framework~\cite{2023mmpretrain}. We utilize the MAE  framework and replace the encoder using \ours while keeping other components unchanged.  This minor modified setting forms a controlled experiment to evaluate the role of our approaches. Moreover, we use the same hyperparameter as \cite{he2022masked}, which is suboptimal to our method. Fortunately, this simple setting still achieves a significant performance boost over the strong baseline.
\par \noindent \textbf{Full fine-tuning.}
In such a setting, the model is first initialized using the pre-trained weights and then trained for extra epochs with totally trainable parameters.  Trained by 800 epochs on the ImageNet, \ours-Base achieves 84.0\% top-1 accuracy, which exceeds ViT-Base by 0.8\%. Note that our method uses a mask ratio of 0.75 as \cite{he2022masked}, whose training speed is about 3 times faster than SimMIM \cite{xie2022simmim}. We also increased the training epochs to 1600 to verify whether \ours keeps the advantage given sufficient training resources. \ours-Base achieves new state-of-art result among MAE variants, 84.3\% top-1 accuracy, which outperforms ViT-Base by 0.9\%. This result is even higher than MaskFeat \cite{wei2022masked} where new training objectives are proposed. Regarding full fine-tuning having a risk of performance saturation \cite{liu2023improving, vishniakov2023convnet}, our boost is significant. Next, we resort to the linear probing metric to provide extra evaluations, which is considered a more reliable evaluation for representative learning by a recent work \cite{chen2024deconstructing}.
\begin{table}
  \setlength\tabcolsep{2pt}
 	\caption{Comparison with MIM SSL methods. $\dagger$: reproduced in MMPretrain.}
  	\label{tab:ssl-imagenet}
	\centering
  \footnotesize
	\begin{tabular}{r*{2}{c}l*{3}{c}}
		\toprule
	     Models&Pretrain Epochs & SFT  Acc(\%)  & LP  Acc(\%)\\

		\midrule
          ViT-Base-MAE$^\dagger$ \cite{he2022masked} & 800 & 83.2& 65.1 \\
          SemMAE \cite{li2022semmae} & 800 & 83.4 & 65.0 \\
          SimMIM \cite{xie2022simmim} &800&83.8&56.7\\
          MFF-MAE \cite{liu2023improving} & 800 & 83.6 & 67.0 \\
\rowcolor{yellow!50}          \ours-Base-MAE & 800 & 84.0& 69.7\\

          \midrule

          ViT-Base-MAE \cite{he2022masked} & 1600 & 83.4& 67.0 \\
          MaskFeat \cite{wei2022masked} & 1600 & 84.0 & 62.3\\
\rowcolor{yellow!50}          \ours-Base-MAE & 1600 &84.3& 71.7\\
          \midrule
          ViT-Large-MAE$^\dagger$ \cite{he2022masked} &800 &85.4&73.7\\
\rowcolor{yellow!50}          \ours-Large-MAE&800 & \textbf{85.5} & \textbf{77.3}\\

		\bottomrule
	\end{tabular}
\end{table}
\par \noindent\textbf{Linear probing.} In this setting, the model is initialized by the pre-trained weights from the  SSL stage. Then, the whole backbone is frozen except for the classifier head during the training. The result is shown in Table~\ref{tab:ssl-imagenet}.  With a training cost of 800 epochs, \ours-Base outperforms ViT-Base-MAE by \textbf{4.6\%}. It also exceeds ViT-Base-MAE, which is trained for 1600 epochs. When \ours is trained for 1600 epochs, \ours-Base achieves 71.7\% top-1 accuracy. We also scale up to have \ours-Large, where our method exceeds ViT-Large by 3.6\%.

\subsection{Semantic Segmentation on ADE20K}
\subsubsection{Supervised Training}
Following \cite{chu2021Twins, liu2021swin}, we evaluate our method using semantic segmentation on the ADE20K~\cite{zhou2017scene} dataset. To make fair comparisons, we limit the baselines to only using ImageNet-1K in the pre-training stage. Specifically, we make use of the UperNet \cite{xiao2018unified} framework and replace the backbone with pyramid \ours.  The detailed setting of the hyperparameter is shown in Section~\ref{hyper:pyramid_ade20k}(supp). We report the result in Table~\ref{tab:ade20k-pyramid}. Our method outperforms both Swin and Twins by more than 1.2\% mIoU.

\begin{table*}[t]
\centering
\caption{\textbf{Segmentation results on ADE20k.} }
\label{tab:ablations} 
\begin{subtable}{0.4\linewidth}
	\centering
	\scriptsize
 	\caption{Performance comparisons with different backbones on ADE20K validation dataset. All backbones are pre-trained on ImageNet-1K with labels.}
  \vspace{.3060cm}
  	\label{tab:ade20k-pyramid}
	\begin{tabular}{rH*{2}{c}l*{3}{c}}
		\toprule
	     Models&FLOPs & Param & mIoU (ss) \\
	     & (G) & (M) & (\%) \\

		\midrule
  		Swin-S \cite{liu2021swin} &261 &81.3&47.6\\

		\altbase \cite{chu2021Twins}  & 261 & 88.5 & 47.7\\
\rowcolor{yellow!50}            \oursp-B &261& 88.5&\textbf{49.1}\\
  
		\midrule
		Swin-B \cite{liu2021swin} & 299& 121& 48.1 \\
    	\altlarge \cite{chu2021Twins} & 297 &133&48.8\\
\rowcolor{yellow!50}             \oursp-L &297&133&\textbf{50.0}\\

		\bottomrule
	\end{tabular}
\end{subtable} 
\hspace{4em}
\begin{subtable}{0.4\linewidth}
 \setlength\tabcolsep{2pt}
	\centering
	\scriptsize 
 	\caption{Performance comparisons with different SSL trained backbones on ADE20K validation dataset.  All models are pre-trained on ImageNet-1K \textbf{ without labels}.$\dagger$: reproduced  using \cite{contributors2020mmsegmentation}.}
	\begin{tabular}{r*{2}H{c}c*{3}{c}}
		\toprule
	     Models&FLOPs & Param& Epochs  & mIoU(ss) \\

      \midrule
      ViT-B$^\dagger$ & &&800&46.2\\
      SemMAE \cite{li2022semmae} &&& 800 &46.3 \\
      MFF-MAE \cite{liu2023improving}&&&800& 47.9 \\

\rowcolor{yellow!50}     \ours-B &&&800&\textbf{49.0}\\
    \midrule
      ViT-B & &&1600&48.1\\
      MaskFeat \cite{wei2022masked} &&&1600& 48.3\\

\rowcolor{yellow!50}       \ours-B &&&1600&\textbf{50.2}\\
		\bottomrule
	\end{tabular}
  \label{tab: ade20k-ssl}
\end{subtable} 
\\
\vspace{-.28cm}
\end{table*}

\subsubsection{Self-Supervised Training}
We use  UperNet \cite{xiao2018unified} to perform semantic segmentation on  ADE20K. We carefully control the experiment and replace the ViT backbone with \ours while keeping other components and hyperparameters unchanged. The detailed hyperparameters are provided in Section~\ref{hyper:ssl_ade20k}(supp.).
The result is given in Table~\ref{tab: ade20k-ssl}. As for the 800 epoch pre-training groups, \ours-B significantly boosts ViT-Base by 2.8\% mIoU. It also outperforms some other modifications such as introducing extra training objectives or features \cite{liu2023improving, wei2022masked} by clear margins. Moreover, those approaches introduce extra overhead for the training process and slow down the training speed. We emphasize that the training speed of a method is becoming more and more important in the age of large models.  In contrast, \ours only involves the replacement of the base model and has the same fast training speed as \cite{he2022masked}. In principle, our method can be seamlessly combined with these modifications. We further evaluate the performance of longer pre-training epochs of 1600, \ours-B achieves 50.2\% mIoU on the ADE20K validation set, which boosts ViT-B by 2.1\% mIoU.

\subsection{Object Detection on COCO}

\subsubsection{Supervised Training.}

We evaluate the performance of pyramid \ours on the COCO objection detection task.  Specifically, we use the Mask RCNN framework \cite{he2017mask} and replace the backbone with pyramid \ours, which is pre-trained for 300 epochs on  ImageNet-1K as \cite{liu2021swin, chu2021Twins}. Since our target is not to achieve a new state-of-the-art detector, this carefully controlled experiment is used to verify the validity of our method.  The hyperparameter setting is provided in Section~\ref{hyper:coco_pyramid}(supp.). We report the result in Table~\ref{tab: pyramid-detection-mask}. Our model outperforms both Swin and Twins. Specifically, \ours-B exceeds Swin-S by 1.5\% box mAP and 1.0 mask mAP. Compared with the stronger baseline Twins-B, ours also has an advantage of 1.1\% higher box mAP and 0.8\% higher mask mAP.

\begin{table}
	\caption{Performance on the COCO2017 dataset 
 using Mask R-CNN.}
    \label{tab: pyramid-detection-mask}
    \setlength\tabcolsep{1pt}
	\centering
	\scriptsize 
	\begin{tabular}{rcH*{7}{c}}
	\toprule
	\multirow{2}{*}{Backbone} &Setting& \multirow{2}{*}{\tabincell{c}{FLOPs (G)}} & \multirow{2}{*}{\tabincell{c}{Param \\(M)}} &\multicolumn{6}{c}{Mask R-CNN 3$\times$ + MS} \\
	&& &&AP$^{\rm b}$ &AP$_{50}^{\rm b}$ &AP$_{75}^{\rm b}$ &AP$^{\rm m}$ &AP$_{50}^{\rm m}$ &AP$_{75}^{\rm m}$ \\
	\midrule
	Swin-S \cite{liu2021swin} & ImageNet1k 300E & 222 & 69.1 & 47.6&69.4& 52.5&42.8&66.5&46.4 \\
	\altbase \cite{chu2021Twins} & ImageNet1k 300E &224&76.3&48.0&69.5&52.7&43.0&66.8&46.6\\
\rowcolor{yellow!50}        Pyramid \ours-B &ImageNet1k 300E &224&76.3 & \textbf{49.1}&\textbf{70.5}& \textbf{54.0}& \textbf{43.8}&\textbf{67.4}& \textbf{47.0}\\
	\bottomrule
\end{tabular}
\end{table}
\subsubsection{Self-Supervised Training.}
 We apply \ours based on ViTDet \cite{li2022exploring}, which utilizes plain transformers to achieve comparable performance as the pyramid counterpart.  Specifically, we replace the vit-Base backbone (trained for 1600 epochs using MAE) with our  \ours-Base model, which is pre-trained for 800 epochs. The original ViTDet converges slowly and requires dedicated training strategies like longer training epochs (\eg, 100) to achieve good performance. During the training process, we find \ours achieves similar performance after 30 epochs. Therefore, we directly utilize the standard 3$\times$ training strategy. Therefore, our training cost is only \textbf{36\%} of the baseline. Unlike \cite{li2022exploring}, we do not search for the optimal hyperparameter. The result is shown in Table~\ref{tab:compare-with-ViTDet} and \ours outperforms ViT-B by 0.6\% Box mAP and 
 0.8\% mask mAP.
\begin{table}[ht]
\caption{Object detection result on COCO 2017 dataset based on  ViTDet\cite{li2022exploring}.}
\label{tab:compare-with-ViTDet}
\centering
\setlength{\tabcolsep}{1pt}
\footnotesize
\begin{tabular}{*{1}{l}*{2}{c}*{10}{c}}
\toprule
Model & Pretrained  & mAP$^{Box}$ & mAP$^{Mask}$ & Epochs\\
\midrule
Swin-S \cite{liu2021swin} &ImageNet sup 300e &47.6 & 42.8 & 36\\
Twins-SVT-B \cite{chu2021Twins} &ImageNet sup 300e&48.0&43.0&36 \\

ViT-B \cite{li2022exploring}  & MAE 1600e &51.6&45.7&100\\
\rowcolor{yellow!50} \ours-B & MAE 800e &\textbf{52.2}&\textbf{46.3}&36\\
\bottomrule
\end{tabular}
\end{table}

\section{Ablation Study and Discussion}
\subsection{Ablation Studies}
Unless otherwise specified, we choose the ViT-Large model (160I 800E+224I 20E) to perform ablations because we observe that it generates small variance across multiple runs, where a performance gap of more than 0.2 suffices as a guide to choosing appropriate components. We give ablations in Table~\ref{tab:ablations} and more ablations are provided in Section~\ref{sec:more_ablations} (supp.).

\begin{table*}[ht]
\scriptsize
\centering
\caption{\textbf{Ablation experiments} with plain transformer ViT-L/16 (DeiT3-L) on ImageNet-1K. We report the top-1 accuracy (\%). If not specified, the default is: and the pre-training length is 800 epochs under an image resolution of 160$\times$160 and 20 epochs using 224$\times$224. Default settings are marked in \colorbox{baselinecolor}{gray}. $\dagger$: running the release code. All accuracies are top-1.}
\label{tab:ablations} 
\begin{subtable}[b]{0.22\linewidth}
\centering
\caption{\textbf{SwiGLU/GELU}}
\label{tab: ablaion_SwiGLU}
\setlength\tabcolsep{2pt}
\begin{tabular}{cc}
case & Acc \\
\hline
\baseline{SwiGLU} & \baseline{84.6}\\
GELU  & 84.6\\
\end{tabular}
\end{subtable} 
\begin{subtable}[b]{0.3\linewidth}
\centering
\caption{\textbf{Normalization}.}
\label{tab: ablation_norm}
\setlength\tabcolsep{2pt}
\begin{tabular}{ccc}
case & Acc & Speed\\
\hline
\baseline{LayerNorm\cite{ba2016layer}} & \baseline{84.6} & 0.4971s \\
RMSNorm \cite{zhang2019root} & 84.4 & 0.4874s\\
\end{tabular}
\end{subtable} 
\begin{subtable}[b]{0.2\linewidth}
\centering
\caption{\textbf{RoPE Ratio}} %
\label{tab: partial ratio}
\setlength\tabcolsep{2pt}
\begin{tabular}{rr}
Ratio & Acc \\
\hline
25\% & 84.5  \\
50\% & 84.5  \\
100\% & \baseline{\textbf{84.6}}  \\
\end{tabular}
\end{subtable} 
\hspace{2em}
\begin{subtable}[b]{0.2\linewidth}
\centering
\caption{\textbf{
RoPE base freq.
} }
\label{tab:abalation_freq_base}
\setlength\tabcolsep{2pt}
\begin{tabular}{rc}
Base & Acc\\
\hline
100 & 84.6 \\
1000& 84.6  \\
10000 & \baseline{\textbf{84.6}}  \\
100000 & 84.4 \\
\end{tabular}
\end{subtable} 
\\
\begin{subtable}[b]{0.25\linewidth}
\centering
\caption{\textbf{Shared PE across different heads}. Shared PE for all heads is better.}

\label{tab:ablation_share_rpe}
\setlength\tabcolsep{2pt}
\begin{tabular}{cc}
Shared PE& Acc \\
\hline
N & 84.2\\
Y & \baseline{\textbf{84.6}}  \\
 \multicolumn{2}{c}{~}\\
\end{tabular}
\end{subtable} 
\begin{subtable}[b]{0.35\linewidth}
\centering
\caption{\textbf{Feature extraction}. The class token is better than GAP in the training setting of DeiT3\cite{touvron2022deit}.}
\label{tab: ablation_feature_extraction}
\setlength\tabcolsep{2pt}
\begin{tabular}{ccc}
Method & Class Head & Acc  \\
\hline
\ours-S & Class Token & 81.6 \\
\ours-S & GAP & 81.8 \\
\midrule
\ours-B & Class Token& 83.6 \\
\ours-B & GAP & 83.6 \\
\midrule
\baseline{\ours}-L  & \baseline{Class Token} & \baseline{84.6}  \\
\ours-L & GAP & 84.3 \\
DeiT3-L \cite{touvron2022deit} & Class Token & 84.5 \\
DeiT3-L$^\dagger$ & GAP & 84.2 \\ 
\end{tabular}
\end{subtable} 
\hspace{.5em}
\begin{subtable}[b]{0.35\linewidth}
\centering
\caption{\textbf{PE comparison}. Applying PEG \cite{chu2023conditional} can further improve the performance. P-V-LLaMA: Pyramid VisionLLaMA, LPE: learnable PE}
\label{tab: ablaion_pe}
\setlength\tabcolsep{2pt}
\begin{tabular}{cc}
case & Acc \\
\hline
\baseline{P-LLaMA-S } & \baseline{81.6}\\
P-V-LLaMA-S + LPE \cite{pmlr-v139-touvron21a} & 81.6 \\
P-V-LLaMA-S + PEG \cite{chu2023conditional}  & \textbf{81.8}\\
\end{tabular}
\end{subtable} 
\end{table*}

\noindent\textbf{Ablation of GELU and SwiGLU}. We replace GELU with SwiGLU and report the result in Table~\ref{tab: ablaion_SwiGLU}. We do not observe performance gaps, therefore, we utilize SwiGLU and avoid introducing extra modifications to the LLaMA architecture. This also motivates us to focus on the ablation of the self-attention block. As we apply multi-head self-attention, the remaining two differences become the normalization and positional encoding. 

\noindent\textbf{Ablation of the normalization strategy.} We compare the two widely used normalization methods in transformers: RMSNorm~\cite{zhang2019root} and LayerNorm~\cite{ba2016layer} and report the result in Table~\ref{tab: ablation_norm}.  The latter has a better final performance, which indicates that \emph{re-centering invariance} is also important in the vision tasks. We also report the training speed by the average time spent per iteration, where LayerNorm is only 2$\%$ slower than RMSNorm. Therefore, we choose LayerNorm instead of RMSNorm for better tradeoff. Note that the training speed might differ across different hardware devices and might also be affected by the overall architecture.

\noindent\textbf{Partial PE.} We adjust the ratio of overall channels using RoPE to report the result in Table \ref{tab: partial ratio}, which shows good performance can be achieved if the ratio is set above a small threshold value. We do not observe significant differences across these settings. Therefore, we keep the default setting of \cite{touvron2023llama} and do not follow  \cite{biderman2023pythia, stablecode2024}.

\noindent\textbf{Positional encoding strategy.} We also add other absolute position encoding strategies such as a learnable PE \cite{pmlr-v139-touvron21a} and PEG \cite{chu2023conditional} on pyramid \ours-S. We use the `small' model due to the existence of a strong baseline and report the result in Table~\ref{tab: ablaion_pe}. While the learnable PE does not boost performance, PEG slightly improves the baseline from 81.6\% to 81.8\%. However, we do not include PEG as a basic component regarding three aspects. Firstly, we try to keep the smallest modifications on LLaMA \cite{touvron2023llama}. Secondly, our target is proposing a universal approach for various tasks like ViT \cite{dosovitskiy2020image}. For masked image frameworks like MAE \cite{he2022masked}, it is non-trivial to keep the reduced training cost of masked tokens if the backbone contains PEG. If we mask patches in the input like \cite{xie2022simmim}, it would greatly slow down the training speed. 

\noindent\textbf{Sensitivity to the input size.} We further compare the performance on the enlarged and commonly used resolutions without training to report the result in Table~\ref{tab: acc_vs_resolution}(supp.). Here we use the pyramid transformer since it is more popular in downstream tasks than the plain counterpart. It is not surprising that 1D-RoPE severely suffers from the changed resolutions. NTK-Aware interpolation with $\alpha=2$ achieves similar performance as the 2D-RoPE, which is indeed NTK-Aware ($\alpha=1$). AS2DRoPE shows the best performance for larger resolution. 
\subsection{Discussion}
We further investigate the underlying mechanisms behind our method’s superior performance over ViT in various tasks. In this section, we discuss the boosted convergence speed and attempt to theoretically rationalize the mechanism.

\textbf{Convergence speed.} For image generation, we study the performance w.r.t the training steps. Specifically, we store the checkpoint at 100k, 200k, 300k, and 400k iterations to calculate the fidelity metrics. Since SDE is significantly slower than ODE, we opt to use the ODE sampler instead. The result of the strictly controlled experiment is listed in Table~\ref{tab: fid_vs_steps}(supp.). It appears that \ours converges much faster than ViT across all models. \oursit with 300k training iterations even outperforms the baseline with 400k steps.

We also compare the convergence speed using the DeiT3-Large under the supervised training setting on ImageNet to show the top-1 validation accuracy during the 800 epochs in Figure~\ref{fig:visionllama-deit3-training-acc}(supp.). It also indicates that \ours converges faster than DeiT3-L. We further compare the training loss across 800 epochs of the ViT-Base model under the MAE framework \cite{he2022masked} and illustrate it in Figure~\ref{fig:visionllama-mae-loss}(supp.). \ours has lower training loss at the beginning and the trend is kept till the end.

\textbf{Theoretical 
analysis. 
} We dive into the mechanism of our positional encodings from the theoretical viewpoint. Without loss of generality, given an input embedding of dimension $d=4$, the query at location $(i,j)$ can be written as  $q_{i,j}$. We use $k_{i,j}$ to represent the key vector at $(i,j)$ and $p_{i,j}$ to be the positional encoding using 2D sin-cos encoding\cite{he2022masked,ma2024sit}. The inner dot product between $q_{i_1,j_1}$ and $k_{i_2,j_2}$ using this additive encoding can be written as, 
\begin{equation}
\begin{split}
        q_{i_1,j_1}^\T k_{i_2,j_2}&=(q_{i_1,j_1}+p_{i_1,j_1})^\T
        (k_{i_2,j_2}+p_{i_2,j_2}) \\   &=q_{i_1,j_1}^\T
k_{i_2,j_2}+p_{i_1,j_1}^\T
p_{i_2,j_2}+q_{i_1,j_1}^\T
p_{i_2,j_2}+p_{i_1,j_1}^\T
k_{i_2,j_2}\\
        &=q_{i_1,j_1}^\T
        k_{i_2,j_2}+f(i_1-i_2,j_1-j_2)+M.
\end{split}
\end{equation}
The first item is the inner dot product of contents. The second item reflects the positional effect in the form of $f(i_1-i_2,j_1-j_2)$, which plays a long-distance decaying effect. However, the third item $M=q_{i_1,j_1}^\T p_{i_2,j_2}+p_{i_1,j_1}^\T k_{i_2,j_2}$ means positions directly interacting with the content features, which slows down the learning process.

In contrast, the inner dot product using RoPE can be written as, 
\begin{equation}
    \begin{split}
({\bf R}_{i_1,j_1}q_{i_1,j_1})^\T
(\textbf{R}_{i_2,j_2}k_{i_2,j_2})&=q_{i_1,j_1}^\T
\textbf{R}^\T_{i_1,j_1}
\textbf{R}_{i_2,j_2}k_{i_2,j_2}\\   &=q_{i_1,j_1}^\T
\textbf{R}_{i_1-i_2,j_1-j_2}k_{i_2,j_2}.
\end{split}
\end{equation}
 $\textbf{R}_{i_1-i_2,j_1-j_2}$ contributes a larger absolute value if the positions of $q$ and $k$ are close, and a smaller value if opposite. This introduces certain localities as a prior bias, which resembles the function of a convolution. Moreover, $\textbf{R}_{i_1-i_2,j_1-j_2}$ adjusts the dot product by the multiplication of a factor between 0 and 1, which is more flexible and faster than the addition of $f(i_1-i_2,j_1-j_2)$. We believe that this flexibility allows the transformer to leverage its model capacity effectively, learning a good representation without dedicating some of that capacity to introducing bias or separating position from content. In this way, \ours not only converges faster but also has better final performance.

\section{Conclusion}
In a nutshell, we present \ours to enjoy the benefits of the LLaMA architecture in the vision modality. It is trained either in supervised or self-supervised schemes to validate the power in a myriad of downstream vision tasks 
including 
image classification, detection, and segmentation, among many others. We particularly explore its image generation capacity under the diffusion framework DiT and SiT to confirm its potency.
We conclude that \ours has a strong potential to serve as a new vision backbone to facilitate a large realm of downstream applications.

\section*{Acknowledgements}

This work was in part supported by
National Key R\&D Program of China (No.\  2022ZD0118700).

\input visionllama-supp.tex

\bibliographystyle{splncs04}

\bibliography{cvml,main}
\end{document}

%% file: visionllama-supp.tex
\appendix

\section{More Experiments}

\subsection{Image Generation}

\textbf{Image generation based on the SiT framework}. SiT\cite{ma2024sit} has a flexible choice of drift and diffusion coefficients, which is supported by the recently proposed interpolant framework \cite{albergo2023stochastic}. It improves the performance of image generation using vision transformers by clear margins. Orthogonally, we replace the vision transformer in SiT with \ours to evaluate the benefits of better model architecture, which we call \oursit. Our implementation is based on the released code of \cite{ma2024sit} with carefully controlled experiments. Specifically, we do not change the hyperparameters, although its default setting may be sub-optimal. All the models are trained using the same number of steps. We use \emph{linear interpolant} and the velocity model for all experiments.  To make fair comparisons, we also rerun the released code and sample 50k 256$\times$256 images using the 250 steps SDE sampler (Euler) and report the result in Table~\ref{tab: sit_sde}. \oursit uniformly outperforms SiT across models with various levels of capacities by clear margins. Compared with SiT-L/2, \oursit-L/2 decreases by 5.0 FID, whose magnitude is larger than the boost from the invention of a new framework (4.0 FID). We also report the more efficient ODE sampler (dopri5) in Table~\ref{tab:compare-with-SiTs}, our performance gap remains. Similar to the observation of \cite{ma2024sit}, SDE has better performance than its ODE counterpart.
\begin{table*}[h]
\centering
\setlength{\tabcolsep}{3pt}
\caption{Image generation of  256$\times$256  using SiT \cite{ma2024sit} without CFG.   The FID is calculated by \textbf{a 250-step SDE} Euler sampler. $\dagger$: reproduced result using the released code.}
\label{tab: sit_sde}
\scriptsize 
\begin{tabular}{*{1}{l}H*{2}{c}*{10}{c}}
\toprule
Model & CFG & Flops & Params & Steps & lr & FID%
$\downarrow$& sFID$\downarrow$ &Precision$\uparrow$ &Recall$\uparrow$ &IS$\uparrow$\\
& & (G) & (M) & (K) & & & & \\
\midrule
SiT-S/2 $^\dagger$ &N & 6.06 &33 & 400 & 0.0001 & 58.15 & 9.12& 41.01&60.23& 24.72\\
\rowcolor{yellow!50} \oursit-S/2 &N & 6.06 &33 & 400 & 0.0001 & 53.90 & 8.78 & 42.98 &60.36& 26.74\\
\midrule
SiT-B/2 $^\dagger$  & N & 23.01 & 130 & 400 & 0.0001& 35.54& 6.57& 52.68 & 64.38&42.33&\\
\rowcolor{yellow!50} \oursit-B/2 &N & 23.01 & 130 & 400& 0.0001&29.53&6.32&56.07&64.07& 50.13\\
\midrule
DiT-L/2 & N & 80.71& 458 &400 & 0.0001 &23.3&-&-&- &- \\

SiT-L/2 $^\dagger$ &N & 80.71& 458 &400 & 0.0001 & 19.34& 5.28&63.00 &63.60&70.47\\
\rowcolor{yellow!50} \oursit-L/2 & N & 80.71& 458 &400 & 0.0001 &14.32& 5.17& 66.39& 63.64& 86.85  \\
\midrule
SiT-XL/2 $^\dagger$  & N & 118.64 & 675 & 400 & 0.0001& 16.98& 5.07&65.12& 64.10&77.06\\
\rowcolor{yellow!50} \oursit-XL/2 &N & 118.64 & 675 & 400 & 0.0001& \textbf{12.20} & 5.03 & 67.86 &63.08& 95.28 \\
\bottomrule
\end{tabular}
\end{table*}

We evaluate the image generation using the 250 steps ODE sampler (dopri5) based on the SiT framework in Table~\ref{tab:compare-with-SiTs}. 
\begin{table*}[h]
\centering
\caption{Image generation comparisons using the SiT framework \cite{ma2024sit} All the models are trained using an image resolution of 256$\times$256 with a global batch size of 256. Metrics are calculated using the sampled 50k images without classifier-free guidance. IS: inception score. The FID is calculated by \textbf{a 250 steps ODE} sampler because of the efficiency, which is a bit different from \cite{ma2024sit}. $\dagger$: reproduced result using the released code.}
\label{tab:compare-with-SiTs}
\setlength{\tabcolsep}{3pt}
\scriptsize 
\begin{tabular}{*{1}{l}H*{2}{c}*{10}{c}}
\toprule
Model & CFG & Flops & Params & Steps & lr & FID%
$\downarrow$& sFID$\downarrow$ &Precision$\uparrow$ &Recall$\uparrow$ &IS$\uparrow$\\
& & (G) & (M) & (K) & & & & \\
\midrule
SiT-S/2 $^\dagger$ &N & 6.06 &33 & 400 & 0.0001 & 59.60 & 9.16& 39.41&58.48& 23.32\\
\rowcolor{yellow!50} \oursit-S/2 &N & 6.06 &33 & 400 & 0.0001 & 54.62 & 8.81 & 41.69 &61.16& 26.07\\
\midrule
SiT-B/2 $^\dagger$  & N & 23.01 & 130 & 400 & 0.0001& 36.90& 6.70& 51.00 & 64.10 &39.78&\\
\rowcolor{yellow!50} \oursit-B/2 &N & 23.01 & 130 & 400& 0.0001&30.23&6.36&54.99&64.90& 48.34\\
\midrule
DiT-L/2 & N & 80.71& 458 &400 & 0.0001 &23.3&-&-&- &- \\
SiT-L/2 $^\dagger$ &N & 80.71& 458 &400 & 0.0001 & 20.14& 5.34&61.53 &64.53&67.08\\
\rowcolor{yellow!50} \oursit-L/2 & N & 80.71& 458 &400 & 0.0001 &14.91& 5.16& 64.97& 64.30& 82.23  \\
\midrule
SiT-XL/2 $^\dagger$  & N & 118.64 & 675 & 400 & 0.0001& 17.83& 5.13&63.52& 64.61&73.64\\
\rowcolor{yellow!50} \oursit-XL/2 &N & 118.64 & 675 & 400 & 0.0001& 12.79 & 5.02 & 66.77 &64.37& 90.93 \\

\bottomrule
\end{tabular}
\end{table*}

\begin{table}
\centering 
\caption{FID calculated with the 250-step ODE sampler in view of efficiency based on the SiT framework.}
\label{tab: fid_vs_steps}
\begin{tabular}{ccccc}
\toprule
Method & 100k & 200k & 300k & 400k\\
\midrule
SiT-S/2 & 89.9&71.9&64.5&59.6\\
\rowcolor{yellow!50} \oursit-S/2&82.88&67.1&59.3&54.6\\
\midrule
    SiT-B/2 & 65.76&48.37&41.05& 36.90 \\
\rowcolor{yellow!50}     \oursit-B/2 & 56.60 & 40.62 &34.09&30.22\\
    \midrule
    SiT-L/2 & 45.07 &29.11 &23.40& 20.14\\
\rowcolor{yellow!50}     \oursit-L/2 & 35.39 &21.82&17.23&14.91\\
    \midrule
    SiT-XL/2 & 42.25& 26.49&20.89&17.83\\
\rowcolor{yellow!50}     \oursit-XL/2 &40.46&19.00&14.84&12.79\\
    \bottomrule
\end{tabular}
\end{table}

\subsection{Image Classification}

\begin{table}
 \caption{Top-1 accuracy comparison on different resolutions. The models are trained on 224 and directly evaluated on other resolutions. }
 \label{fig: acc_vs_resolution}
 \setlength\tabcolsep{4pt}
	\begin{center}
		\begin{tabular}{lcccccccc}
		\toprule
			Model & 160 & 224 &  256 & 288 & 512 & 768  \\
            \midrule
                DeiT3-Large \cite{touvron2022deit} & 83.1&84.5&84.7& 84.6& 82.1& 76.5\\
\rowcolor{yellow!50}                 \ours-L & 83.1  & \textbf{84.6} &84.7 &\textbf{84.8} & \textbf{83.5} & \textbf{79.1}\\
            \midrule
		\end{tabular}
	\end{center}
\end{table}

\begin{figure}[t]
	\centering
	\includegraphics[width=0.805\columnwidth]{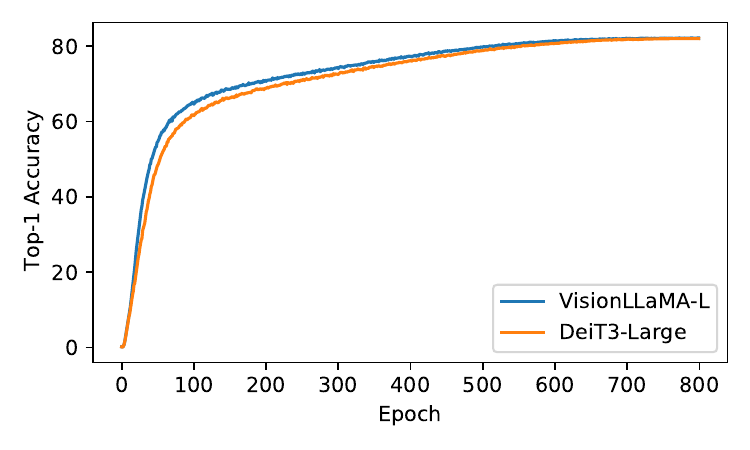}
        \vskip -0.1in
	\caption{Faster convergence of \ours using the setting of DeiT3.}
	\label{fig:visionllama-deit3-training-acc}
\end{figure}

\begin{figure}[ht]
	\centering
	\includegraphics[width=0.805\columnwidth]{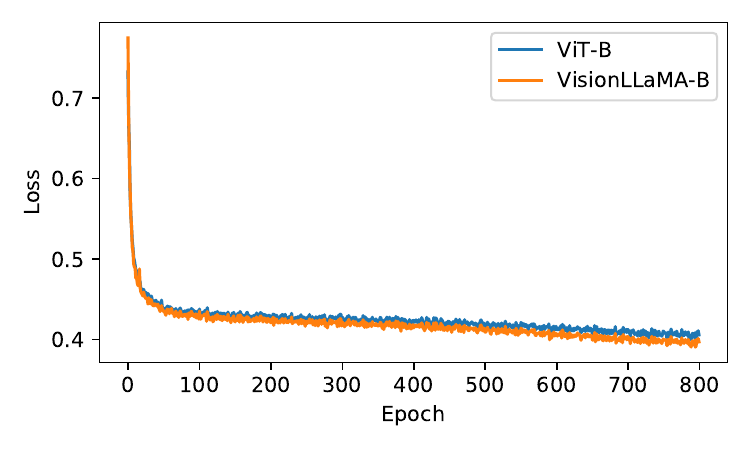}
        \vskip -0.1in
	\caption{Loss curve of MAE pre-training on \ours compared with ViT-B.}
	\label{fig:visionllama-mae-loss}
\end{figure}

\begin{figure}[ht]
	\centering
	\includegraphics[width=0.5\columnwidth]{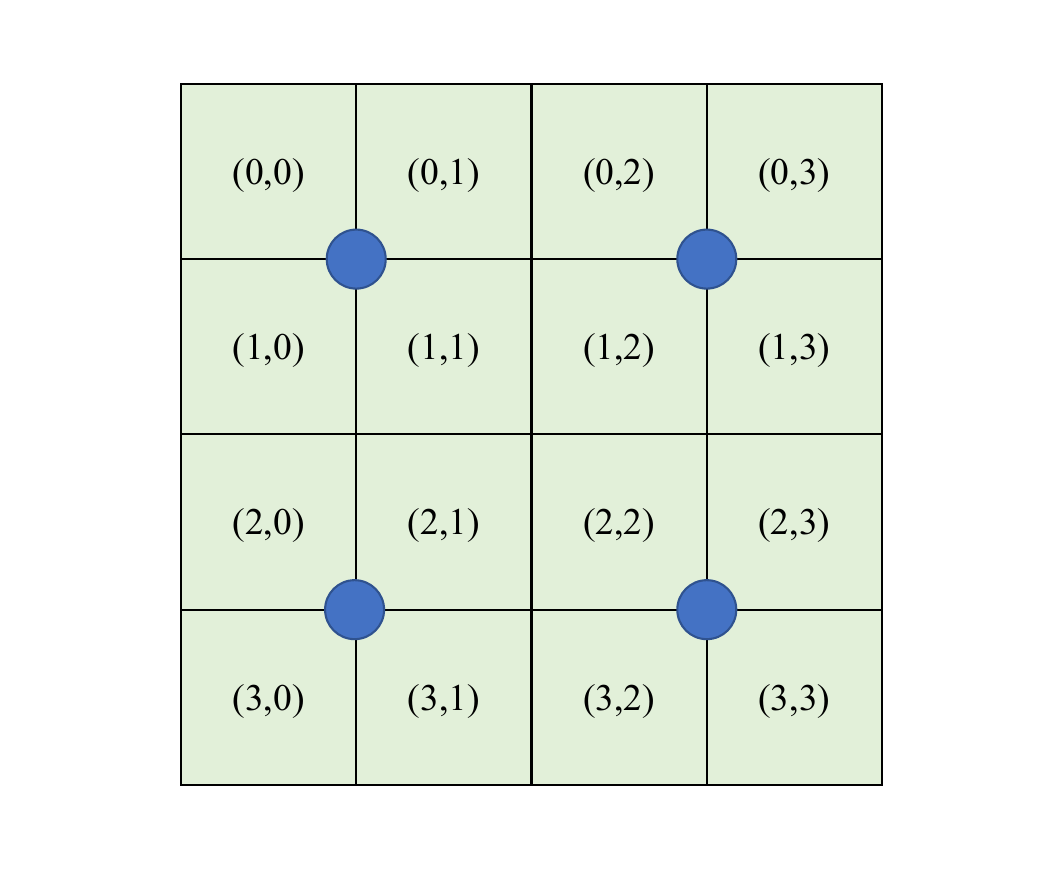}
        \vskip -0.1in
	\caption{Position calibration for GSA's keys using a simple case of $4\times4$ resolution and a kernel size of $2\times2$. The positions of the four points (abstraction keys) are (0.5, 0.5), (1, 2.5), (2.5, 0.5), (2.5, 2.5). }
	\label{fig:gsa-encoding}
\end{figure}

\section{More Ablations}\label{sec:more_ablations}

\noindent\textbf{Shared PE for each head.}  We find that sharing the same PE across different heads (the frequency varies from 1 to 10000 in each head) is better than independent ones (the frequency varies from 1 to 10000 across all channels). The result is shown in Table~\ref{tab:ablation_share_rpe}.

\noindent\textbf{Frequency base.} We change the base frequency and report the result in Table~\ref{tab:abalation_freq_base}, which means the performance is robust to a large range of frequencies. As a result, we keep the default value of \cite{touvron2023llama} to avoid extra special treatments for deployment.

\noindent\textbf{Feature abstraction strategy}. We compare the two common feature extraction strategies: class token \cite{dosovitskiy2020image} and  GAP \cite{chu2023conditional} using the plain `large' model and report the result in Table~\ref{tab: ablation_feature_extraction}. Using a class token is better than GAP, which is different from \cite{chu2023conditional}.  However, the training settings of the two cases are quite different. We also make an extra experiment using DeiT3-L to observe a similar performance gap of 0.3\%. We further evaluate the performance of the `small' and `base' models. It's interesting to see the opposite conclusions for the small model. We suspect that the higher drop-path rate used in \cite{touvron2022deit} makes it difficult for the parameter-free abstraction such as GAP to fit in the purpose.

\begin{table}
 \setlength\tabcolsep{4pt}
 \caption{Top-1 accuracy on different resolutions of the pyramid small model. The models are trained on 224$\times$224 and directly evaluated on other resolutions. }
 \label{tab: acc_vs_resolution}
	\begin{center}
		\begin{tabular}{ lcccccccc }
		\toprule
			Model & 224 &  448 & 512   \\
            \midrule
                1D-RoPE & 81.5& 0.01 & 0.01 \\
                2D-RoPE  & 81.6& 79.5& 78.4\\
                NTK($\alpha=2$)&81.6&79.6&78.5\\
                NTK($\alpha=5$)& 81.3&79.6&78.6\\
                NTK($\alpha=10$)& 81.1  &79.6 & 78.6\\
\rowcolor{yellow!50}                 AS2DRoPE & 81.6& 80.3& 79.5&\\
            \midrule
		\end{tabular}
	\end{center}
\end{table}

\section{More Related Work}
 \par \noindent \textbf{Masked Image Modeling.} 
 Masked image modeling is a powerful pre-training scheme that learns strong representations.
BEiT~\cite{bao2021beit} extends BERT~\cite{devlin2018bert} to computer vision by pre-training a Transformer model with masked embeddings to predict discrete visual tokens. 
Masked Autoencoder (MAE) ~\cite{he2022masked} is a self-supervised learning approach that masks random patches of input images and trains an autoencoder to reconstruct the original images.
SiMMIM ~\cite{xie2022simmim} is a simplified version of the MAE approach that uses a lightweight one-layer head to predict raw pixel values.
MaskFeat~\cite{wei2022masked} is an extension of the MAE approach that involves predicting not only the raw pixel values of the masked patches but also additional features such as handcrafted HOG descriptor~\cite{dalal2005histograms} and deep features, which can improve the performance of the model on downstream tasks.

\par \noindent \textbf{Diffusion Models.} Diffusion models, represented by Denoising Diffusion Probabilistic Models (DDPMs) ~\cite{ho2020denoising,sohl2015deep}, score-based generative models (SGMs) ~\cite{Hyvarinen2005,Song2020b} and classifier-free diffusion guidance~\cite{HoSalimans2021}, are the new de facto paradigm for image generation, surpassing the previous methodology  GAN~\cite{Goodfellow2014}. The mechanism of diffusion models is based on the idea of gradually adding noise to data and then learning to denoise it. Challenges remain for the computationally expensive training and sampling process, 
the need for large amounts of data for training, and the difficulty in controlling the generation process.
Most lately, OpenAI brings about transformer-based text-conditional diffusion models (the largest one called Sora) ~\cite{videoworldsimulators2024} jointly trained on videos and images of variable durations, resolutions, and aspect ratios to deliver high-fidelity videos simulating real-world scenes. The recent and concurrent work\cite{lu2024fit} explores how to deal with image generation with flexible target resolutions. Compared with \cite{lu2024fit}, our target is to build a universal vision transformer for various vision tasks.

\section{Hyperparameters}
\subsection{Supervised Training of \ours on ImageNet-1K}\label{sec:hyper_plain}
As for the plain transformer, we use the same hyperparameters as \cite{touvron2022deit}. The detailed setting is provided in Table~\ref{tab: hyper_image1k_sft}. \ours-S is trained on ImageNet-1K for 800 epochs with a resolution of 224$\times$224. \ours-B is first trained for 800 epochs with an input size of 192$\times$ 192 and then fine-tuned for 20 epochs on $224\times224$. \ours-L is first trained for 800 epochs with a resolution of $160\times160$ and then finetuned for 20 epochs on $224\times224$.

\begin{table}[ht]

\caption{
Training procedures with ImageNet-1K. 
\label{tab: hyper_image1k_sft}}
\centering
\scalebox{1}
{%
\begin{tabular}{@{\ }l|HHcH|cHH@{\ }}
\toprule
 & \multicolumn{4}{c|}{Sec~\ref{sec:hyper_pyramid}} &  \multicolumn{3}{c}{Sec~\ref{sec:hyper_plain}} \\

\midrule
Batch size & 
4096 & 
4096 &
1024 & 
2048 & 
2048 & 2048 & 2048\\
Optimizer &
AdamW & 
AdamW &
AdamW &
LAMB &
LAMB &
LAMB & LAMB\\
LR      & 
$3.10^{-3}$ & 
$3.10^{-3}$ &
$1.10^{-3}$  & 
$5.10^{-3}$  & 
$3.10^{-3}$ &
$3.10^{-3}$ & 
$3.10^{-4}$\\
LR decay& 
cosine  &
cosine & 
cosine & 
cosine & 
cosine &
cosine & cosine\\
Weight decay     &
0.1  & 
0.3  & 
0.05 & 
0.02 &
0.02 &
0.02 & 0.02\\
Warmup epochs & 
3.4 &
3.4 &
5 & 
5  &
5 &
5 & 5 \\
\midrule
Label smoothing $\varepsilon$ & 
0.1 & 
0.1 &
0.1  &
\xmarkg & 
\xmarkg  &
0.1  & 0.1 \\%
Dropout      & 
\cmark  & 
\cmark & 
\xmarkg & 
\xmarkg  & 
\xmarkg &
\xmarkg & \xmarkg\\
Stoch. Depth & 
\xmarkg & 
\cmark & 
\cmark & 
\cmark &
\cmark&
\cmark & \cmark \\
Repeated Aug & 
\xmarkg & 
\xmarkg & 
\cmark &
\cmark &
\cmark &
\xmarkg & \xmarkg\\
Gradient Clip. & 
1.0  & 
1.0 & 
\xmarkg & 
1.0 & 
1.0 &
1.0 & 1.0\\
\midrule
H. flip  & 
\cmark & 
\cmark &
\cmark & 
\cmark & 
\cmark  &
\cmark  & \cmark
\\
RRC & 
\cmark & 
\cmark & 
\cmark & 
\cmark & 
\cmark & 
\xmarkg & \xmarkg\\
Rand Augment  &
\xmarkg & 
Adapt. &
9/0.5 &
7/0.5 &
\xmarkg &
\xmarkg & \xmarkg  \\
3 Augment   &
\xmarkg  & 
\xmarkg  &
\xmarkg  &
\xmarkg &
\cmark &
\cmark & \cmark \\

LayerScale  & 
\xmarkg  & 
\xmarkg  & 
\xmarkg  &
\xmarkg  &
  \cmark&
  \cmark &  \cmark\\

Mixup alpha  & 
\xmarkg & 
Adapt. & 
0.8 &
0.2 & 
0.8  &
\xmarkg & \xmarkg \\
Cutmix alpha &
\xmarkg & 
\xmarkg & 
1.0 &
1.0 & 
1.0 &
1.0 & 1.0 \\
Erasing prob. &
\xmarkg    &
\xmarkg    &
0.25 &
\xmarkg  &
\xmarkg & 
\xmarkg & \xmarkg \\
ColorJitter  & 
\xmarkg   & 
\xmarkg   & 
\xmarkg  &
\xmarkg  &
  0.3 & 
 0.3 & 0.3\\

\midrule
Test  crop ratio & 
0.875 & 
0.875 &
0.875 & 
0.95 &
1.0 & 
1.0 & 1.0 \\
\midrule
Loss &
CE & 
CE & 
CE &
BCE & 
BCE &
CE & CE \\

 \bottomrule
\end{tabular}}
\end{table}

\subsection{Supervised Training of Pyramid \ours}\label{sec:hyper_pyramid}
We use the same setting as \cite{chu2021Twins}. Specifically,  all the models are trained on ImageNet-1K for 300 epochs with a global batch size of 1024 using the AdamW optimizer. The learning rate is increased to 0.001 within 5 warm-up epochs and decayed to zero following the cosine schedule. We use the same data augmentation as \cite{chu2021Twins} and an image resolution of 224$\times$224 for all models. To avoid overfitting, we use a weight decay of 0.05 and drop path \cite{huang2016deep} (0.2, 0.3, 0.5 for small base and large models respectively).  
\subsection{Mask Image Modeling on ImageNet}
We use AdamW optimizer with momentum $\beta_1=0.9$ and $\beta_2=0.95$. The global batch size is 4096. The initial learning rate is 1.5$\times$10$^-4$ and decayed to zero within 800 or 1600 epochs. We also use 40 epochs to warm up the learning rate. 
We only use simple data augmentation RRC(random-resize-crops)as \cite{he2022masked}. Besides, we use a weight decay of 0.05.
\subsection{Linear Probing on ImageNet}
We follow the setting of \cite{he2022masked} and show the details in Table~\ref{tab:hyper_mae_linear}.

\begin{table}[ht]
\centering
\caption{\textbf{Linear probing setting.}
\label{tab:hyper_mae_linear}}
\begin{tabular}{ r c}
\centering
config & value \\
\shline
optimizer & LARS \cite{you2017large} \\
base learning rate & 0.1 \\
weight decay & 0 \\
optimizer momentum & 0.9 \\
batch size & 16384 \\
learning rate schedule & cosine decay \\
warm-up epochs & 10 \\
training epochs & 90 \\
augmentation & RandomResizedCrop \\
\bottomrule
\end{tabular}
\end{table}

\subsection{SFT on ImageNet for SSL-pre-trained Models}
We follow the same setting of \cite{he2022masked} and show the details in Table~\ref{tab:mae_finetune}. The only modification is the layer-wise learning rate decay because we find  0.75 of \cite{he2022masked} is overfitting for our method, and we set it to 0.45. 
\begin{table}[ht]
\centering
\caption{\textbf{End-to-end fine-tuning setting for SSL.}}
\label{tab:mae_finetune} 
\begin{tabular}{ r c}
config & value \\
\shline
optimizer & AdamW \\
base learning rate & 1e-3 \\
weight decay & 0.05 \\
optimizer momentum & $\beta_1, \beta_2{=}0.9, 0.999$ \\
layer-wise lr decay \cite{bao2021beit} & 0.45 \\
batch size & 1024 \\
learning rate schedule & cosine decay \\
warmup epochs & 5 \\
training epochs & 100 (B), 50 (L) \\
augmentation & RandAug (9, 0.5) \cite{cubuk2020randaugment} \\
label smoothing \cite{szegedy2016rethinking} & 0.1 \\
mixup \cite{zhang2017mixup} & 0.8 \\
cutmix \cite{yun2019cutmix} & 1.0 \\
drop path & 0.1  \\
\bottomrule
\end{tabular}
\end{table}

\subsection{Segmentation for SSL-pre-trained Models} \label{hyper:ssl_ade20k}
We follow the default setting of \cite{chu2021Twins}. We use AdamW \cite{loshchilov2017decoupled}  optimizer with $\beta_1=0.9$ and $ \beta_2=0.999$. The global batch size is 16. The initial learning rate is 6$\times$ $10^{-5}$ and linearly decayed to zero. We also use 1500 iterations to warm up. We also utilize $l_2$ weight decay of 0.05 and a drop-path rate \cite{huang2016deep} of 0.1.

\subsection{Segmentation for Pyramid Transformers} \label{hyper:pyramid_ade20k}
We follow the setting of \cite{chu2021Twins}, which is almost the same as \ref{hyper:ssl_ade20k}. We use a drop-path rate of 0.2 for the pyramid \ours-B model.

\subsection{Object Detection of Pyramid Transformers} \label{hyper:coco_pyramid}
We use the same setting as \cite{chu2021Twins}. We use the AdamW optimizer with $\beta_1=0.9$ and $\beta_2=0.999$. All the models are trained for 36 epochs with a global batch size of 16. The initial learning rate is 1$\times$10$^{-4}$ with 1000 iterations warm up and decayed by 10.0 at epoch 27 and 33. To avoid overfitting, we apply a $l_2$ weight decay for all models.

\subsection{Object Detection of Plain Transformers}
 We use the AdamW optimizer with $\beta_1=0.9$ and $\beta_2=0.999$. The training resolution is fixed as 1024 $\times$ 1024 as \cite{li2022exploring}. Our model is trained for 36 epochs with a global batch size of 64. The initial learning rate is 1$\times$10$^{-4}$ with 1000 iterations warm up and decayed by 10.0 at epoch 27 and 33. We use a $L_2$ weight decay of 0.1. We also apply layer-wise learning rate decay with 0.7 as \cite{li2022exploring}.

\subsection{Image Generation of \ourdit}
We use the same VAE as \cite{rombach2022high}. We make use of the AdamW optimizer with momentum $\beta_1=0.9$ and $\beta_2=0.999$. We use a global batch size of 256 across all models. The learning rate is fixed as 1 $\times$ $10^{-4}$. The training resolution is 256 $\times$ 256. As for inference, we use 250 steps DDPM. We keep the default setting of ADM \cite{dhariwal2021diffusion} without tuning. Specifically, we use $t_{\max}=1000$ linear schedule with $\beta$ from 0.0001 to 0.02 and learnable variance $\sigma_\theta$.
\subsection{Image Generation of \oursit}
We use the same VAE as SD \cite{rombach2022high}. As for the ODE sampler, we utilize dopri5 and set `atol' and `rtol' to 1e$-6$ and 1e$-3$ respectively.

\section{Architecture Setting}

\subsection{Pyramid \ours}

The detailed setting of the pyramid architecture is shown in Table~\ref{tab:pyramid_arch}.

\begin{table*}[ht]
	\tabcolsep 1pt
	\caption{%
		Configuration details of Pyramid \ours.}
 \label{tab:pyramid_arch}
	\scriptsize
	\centering 
	\begin{tabular}{*{5}{c|}cc}
		\toprule
		& Output Size & Layer Name &S & B  & L &   \\
		\midrule
		\multirow{2}{*}[-2.5ex]{Stage 1} & \multirow{2}{*}[-2.5ex]{\scalebox{1.3}{$\frac{H}{4}\times \frac{W}{4}$}} & Patch Embedding &$P_1=4$;  $C_1=64$ &$P_1=4$; $C_1=96$ & $P_1=4$; $C_1=128$\\
		\cline{3-6}
		& &  & 
		$\begin{bmatrix}
			\begin{array}{l}
				LSA \\
				GSA \\
				
			\end{array}
		\end{bmatrix} \times 1$ &
		$\begin{bmatrix}
			\begin{array}{l}
				LSA \\
				GSA \\
			\end{array}
		\end{bmatrix} \times 1$ &
		$\begin{bmatrix}
			\begin{array}{l}
				LSA \\
				GSA \\
			\end{array}
		\end{bmatrix} \times 1$ \\
		\midrule
		\multirow{2}{*}[-2.5ex]{Stage 2} & \multirow{2}{*}[-2.5ex]{\scalebox{1.3}{$\frac{H}{8}\times \frac{W}{8}$}} & Patch Embedding & $P_2=2$;  $C_2=128$&$P_2=2$;  $C_2=192$&$P_2=2$;  $C_2=256$  \\
		\cline{3-6}
		& &  &
		$\begin{bmatrix}
			\begin{array}{l}
				LSA\\
				GSA\\
			\end{array}
		\end{bmatrix} \times 1$ &
		$\begin{bmatrix}
			\begin{array}{l}
				LSA\\
				GSA\\
			\end{array}
		\end{bmatrix} \times 1$ &
		$\begin{bmatrix}
			\begin{array}{l}
				LSA\\
				GSA\\
			\end{array}
		\end{bmatrix} \times 1$\\
		\midrule
		\multirow{2}{*}[-2.5ex]{Stage 3} & \multirow{2}{*}[-2.5ex]{\scalebox{1.3}{$\frac{H}{16}\times \frac{W}{16}$}} & Patch Embedding & $P_3=2$;  $C_3=256$& $P_3=2$;  $C_3=384$& $P_3=2$;  $C_3=512$\\
		\cline{3-6}
		& &  &
		$\begin{bmatrix}
			\begin{array}{l}
				LSA\\
				GSA \\
			\end{array}
		\end{bmatrix} \times 5$ &
		$\begin{bmatrix}
			\begin{array}{l}
				LSA\\
				GSA \\
			\end{array}
		\end{bmatrix} \times 9$ &
		$\begin{bmatrix}
			\begin{array}{l}
				LSA\\
				GSA \\
			\end{array}
		\end{bmatrix} \times 9$\\
		\midrule
		\multirow{2}{*}[-1ex]{Stage 4} &  \multirow{2}{*}[-1ex]{\scalebox{1.3}{$\frac{H}{32}\times \frac{W}{32}$}} & Patch Embedding &$P_4=2$; $C_4\!=\!512$ & $P_4=2$; $C_4\!=\!768$&$P_4=2$; $C_4\!=\!1024$\\
		\cline{3-6}
		& &  &
		$\begin{bmatrix}
			\begin{array}{l}
				GSA\\
			\end{array}
		\end{bmatrix} \times 4$ & $\begin{bmatrix}
			\begin{array}{l}
				GSA\\
			\end{array}
		\end{bmatrix} \times 2$ & $\begin{bmatrix}
			\begin{array}{l}
				GSA\\
			\end{array}
		\end{bmatrix} \times 2$\\
		\bottomrule
	\end{tabular}
\end{table*}

\subsection{Plain Transformer for Vision Understanding}\label{sec:plain_arch_understanding}
The detailed setting of the architecture is shown in Table~\ref{tab:plain_arch_understanding}.

\begin{table}[]
    \centering
    \caption{Architecture settings for \ours on image understanding tasks.}
    \label{tab:plain_arch_understanding}
    \begin{tabular}{cccc}
    Model & Layers & Dims & Heads \\
    \toprule
         \ours-S & 12&384& 6  \\
         \ours-B &12 & 768&12 \\
         \ours-L & 24 & 1024&16 \\
         \bottomrule
    \end{tabular}
\end{table}

\subsection{Plain Transformer for Generation}\label{sec:plain_arch}
The detailed setting of the architecture is shown in Table~\ref{tab:plain_arch}.
\begin{table}[]
    \centering
    \caption{Architecture settings for \ours on image generation.}
    \label{tab:plain_arch}
    \begin{tabular}{cccc}
    Model & Layers & Dims & Heads \\
    \toprule
         \ourdit-S / \oursit-S& 12&384& 6  \\
         \oursit-B / \oursit-B &12 & 768&12 \\
         \oursit-L / \oursit-L & 24 & 1024&16 \\
         \oursit-XL / \oursit-XL & 28 & 1152&16\\
         \bottomrule
    \end{tabular}
\end{table}%